%% file: main.tex
\documentclass{article}

\PassOptionsToPackage{numbers, compress}{natbib}

\usepackage[preprint]{neurips_2025}

\usepackage[utf8]{inputenc} 
\usepackage[T1]{fontenc}    
\usepackage{hyperref}       
\usepackage{url}            
\usepackage{booktabs}       
\usepackage{amsfonts}       
\usepackage{nicefrac}       
\usepackage{microtype}      
\usepackage{xcolor}         

\usepackage{comment}
\usepackage{graphicx}
\title{Single-Stage Huffman Encoder for ML Compression}

\author{%
  Aditya Agrawal \\
  \texttt{adityaag@google.com} \\
  \And
  Albert Magyar \\
  \texttt{amagyar@google.com} \\
  \And
  Hiteshwar Eswaraiah \\
  \texttt{hiteshwarbe@google.com} \\
  \And
  Patrick Sheridan \\
  \texttt{sheridp@google.com} \\
  \And
  Pradeep Janedula \\
  \texttt{pjanedula@google.com} \\
  \And
  Ravi Krishnan Venkatesan \\
  \texttt{ravikrishnan@google.com} \\
  \And
  Krishna Nair \\
  \texttt{krishnaknair@google.com} \\
  \And
  Ravi Iyer \\
  \texttt{raviyer@google.com} \\
  \AND
  Google LLC
}

\begin{document}

\maketitle

\begin{abstract}
\input{abstract}
\end{abstract}

\section{Background}
\input{background}

\section{Experimental Setup}
\input{setup}

\section{Results}
\input{results}

\section{Implementation}
\input{implementation}

\section{Conclusion}
\input{conclusion}

\bibliography{citation}

\end{document}

%% file: abstract.tex
Training and serving Large Language Models (LLMs) require partitioning data across multiple accelerators, where collective operations are frequently bottlenecked by network bandwidth. Lossless compression using Huffman codes is an effective way to alleviate the issue,
however, its three-stage design requiring on-the-fly frequency analysis, codebook generation
and transmission of codebook along with data introduces computational, latency and data overheads which are prohibitive for latency-sensitive scenarios such as die-to-die communication.
This paper proposes a single-stage Huffman encoder that eliminates these overheads
by using fixed codebooks derived from the average probability distribution of previous data batches. Through our analysis of the Gemma 2B model, we demonstrate that tensors exhibit high
statistical similarity across layers and shards. Using this approach we achieve compression
within 0.5\% of per-shard Huffman coding and within 1\% of the ideal Shannon compressibility, enabling efficient on-the-fly compression.

%% file: background.tex
Training and serving Large Language Models (LLMs) e.g.,
Gemini \cite{models:gemini_2025}, Gemma \cite{models:gemma1, models:gemma2}, LLaMA \cite{models:llama}, GPT \cite{models:gpt_1}
require partitioning (sharding) the data (parameters, activations, optimizer state etc.) and parallelizing the computation across multiple accelerators.
There are multiple paradigms of parallelism e.g., Data Parallelism, Tensor Parallelism, Pipeline Parallelism, Expert Parallelism and Sequence Parallelism \cite{parallelism:megatron, parallelism:sequence, parallelism:zero, parallelism:nvidia_1, parallelism:nvidia_2, parallelism:colossalai, parallelism:scaling-book}.
Different parallelization strategies invoke different collective operations e.g., AllReduce, ReduceScatter, AllGather, AlltoAll \cite{collectives:nccl}. Collective operations are typically bounded by network bandwidth.


Lossless compression is an effective way to reduce the network traffic and improve collective performance. Huffman codes \cite{huffman} either directly or as part of other algorithms e.g., DEFLATE \cite{deflate}, Zstandard \cite{zstd}, Brotli \cite{brotli} are commonly used for lossless data compression. They exploit the distribution of symbol frequencies and are optimal entropy codes.

A Huffman encoder typically has three stages. In the first stage we scan the entire input to build a frequency table of the symbols. In the second stage, we run the Huffman algorithm to generate the codes, typically variable length, for each symbol. Finally, we scan the input again and replace the symbols with their corresponding codes.

The three-stage encoder is useful when the overhead of compression is compensated by a reduction in the network transfer time. However, for extremely latency sensitive scenarios e.g., die-to-die communication, building a frequency table and running the Huffman algorithm is computationally expensive and adds significant latency. In addition, the code book used for encoding has to be communicated to the receiver. These computation, latency and data overheads can erode any benefits of doing on-the-fly compression.

%% file: setup.tex
We analyzed the Gemma 2B model \cite{models:gemma1} during Supervised Fine Tuning (SFT). The model has 18 layers and is sharded over 64 TPUs. We analyzed the weight, activation, weight gradient and activation gradient tensors of the feed forward layers, FFN1 and FFN2. Overall, there are $18 \times 64 = 1152$ shards of each tensor type e.g., FFN1 activation. We analyzed the compressibility at different data types, namely, bfloat16 \cite{format:bfloat16}, e4m3, e3m2, e2m3 and e2m1 \cite{format:exmy, format:ocpmx}. We present our observations and compressibility results for FFN1 activation tensor at data type bfloat16.

%% file: results.tex
\begin{figure}[h]
  \centering
  \includegraphics[width=\linewidth]{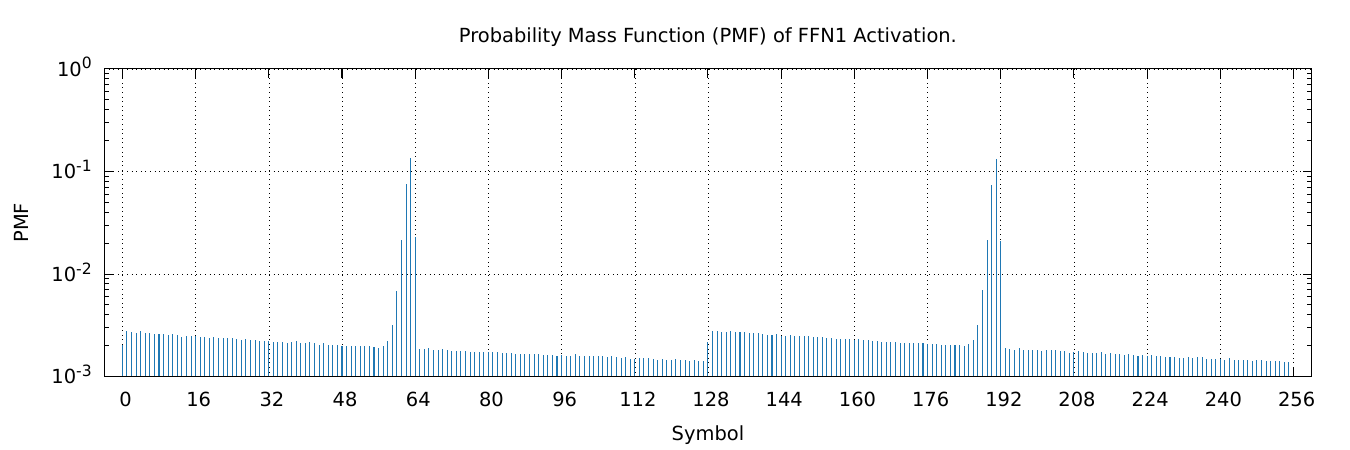}
  \caption{Probability Mass Function (PMF) of FFN1 activation.}
  \label{fig:pmf_ffn1_activation}
\end{figure}
\begin{figure}[h]
  \centering
  \includegraphics[width=\linewidth]{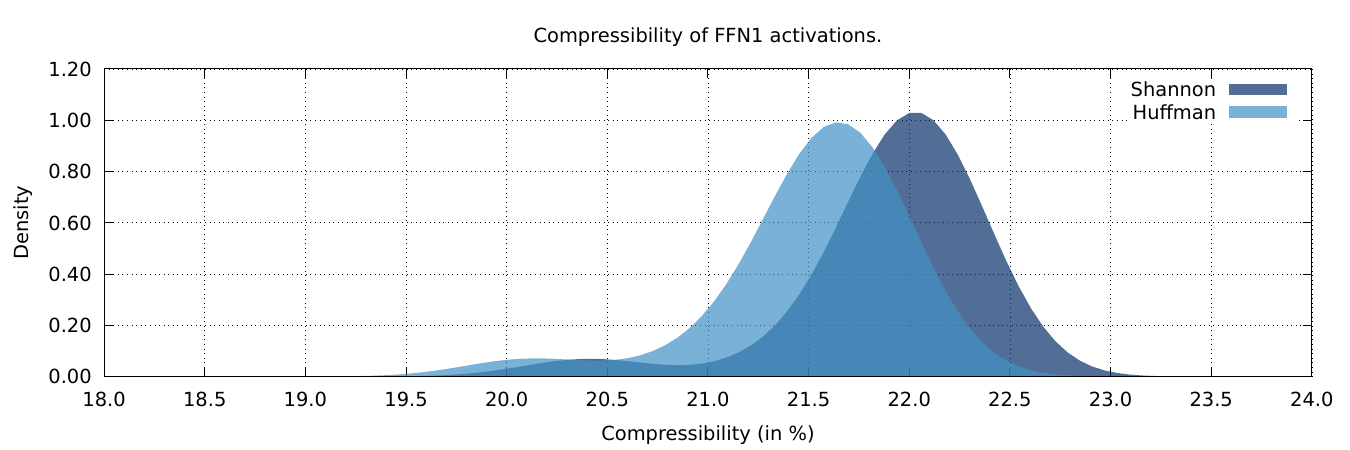}
  \caption{Compressibility of FFN1 activation shards using Huffman codes.}
  \label{fig:shannon_huffman_ffn1_activation}
\end{figure}

Fig. \ref{fig:pmf_ffn1_activation} shows the Probability Mass Function (PMF) of one shard of FFN1 activation with a symbol size of 8 bits i.e., 256 symbols. This distribution has a Shannon entropy \cite{entropy} of 6.25 bits and hence an ideal compressibility of $\frac{8-6.25}{8} \approx 21.9\%$. For this distribution, the compressibility achieved using Huffman codes is $\approx 21.6\%$.
Fig. \ref{fig:shannon_huffman_ffn1_activation} shows the distribution of the ideal per shard compressibility and the compressibility achieved using per shard Huffman codes for all 1152 shards.
The ideal compressibility of most shards is $\approx 21 - 23\%$.
As expected, Huffman codes achieve close to the ideal compression, however, this requires a three-stage encoder.

\begin{figure}[h]
  \centering
  \includegraphics[width=\linewidth]{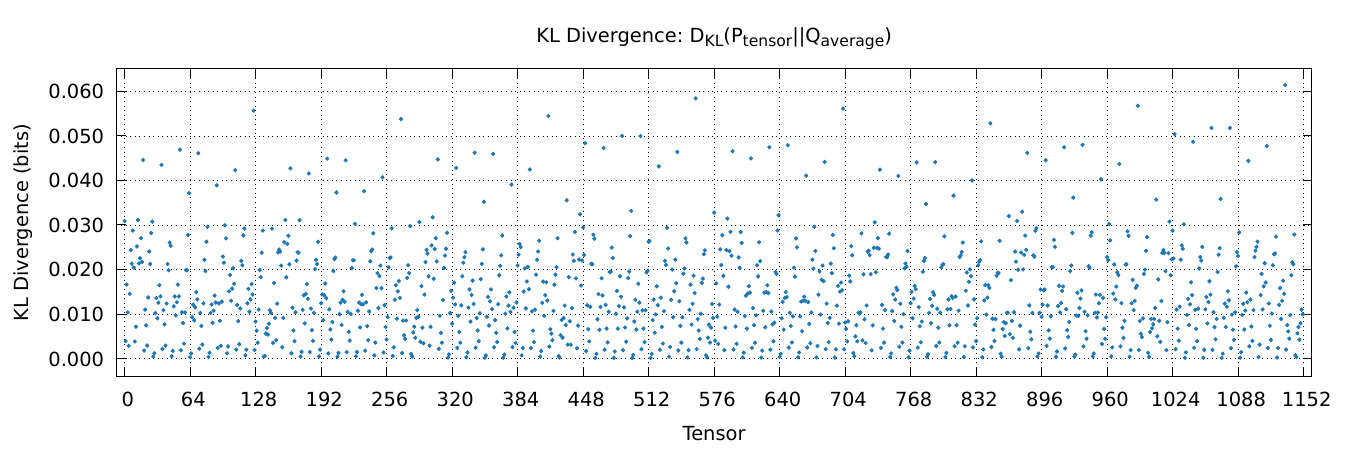}
  \caption{KL divergence of FFN1 activation shards from the average PMF.}
  \label{fig:kl_div_ffn1_activation}
\end{figure}

We observed that the PMF of all the FFN1 activation shards were very similar. Instead of calculating the KL divergence \cite{kl_divergence} for all $1152^2$ shard pairs, we obtained the average PMF and then calculated the KL divergence of each shard from this average distribution. Fig. \ref{fig:kl_div_ffn1_activation} shows the KL divergence of each shard from the average distribution. A small KL divergence (< 0.06) confirms that the PMF of the different shards are indeed similar and that the average distribution is a good approximation of the true distribution.

\begin{figure}[h]
  \centering
  \includegraphics[width=\linewidth]{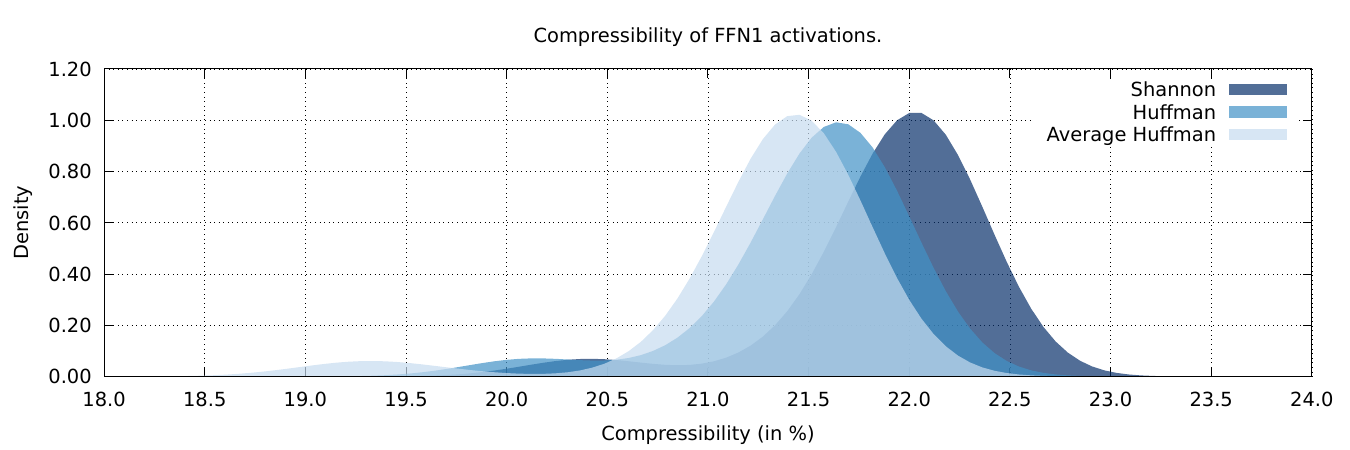}
  \caption{Compressibility using Huffman codes derived from the average distribution.}
  \label{fig:shannon_huffman_avg_ffn1_activation}
\end{figure}

Fig. \ref{fig:shannon_huffman_avg_ffn1_activation} shows the distribution of ideal per shard compressibility, compressibility achieved using per shard Huffman codes and compressibility achieved using Huffman codes obtained from the average distribution and then applied to all shards.
Huffman codes obtained from the average distribution achieve a compressibility within 1\% of the ideal Shannon compressibility and within 0.5\% of the compressibility achieved by using per shard Huffman codes.

The histograms and compressibility are different for other tensors and datatypes, however, they still exhibit statistical similarity between shards and codebooks derived from the average distribution achieve compression close to that achieved using per shard Huffman codes.

%% file: implementation.tex
The average distribution can be obtained from previous batches during training or serving. The Huffman codes can be obtained from the average distribution off the critical path. A system of accelerators for ML training, fine tuning, and serving will have multiple code books, one for each tensor e.g., FFN1 activation, FFN2 weight gradient etc.

The selection of a specific code book can be done in software or hardware. In a software implementation, the code book is selected by the programmer. In a hardware implementation, multiple code books can be evaluated for compressibility in parallel. The code book which achieves the best compression is selected.
The code books are shared between the participating nodes and so the encoder sends only the encoded values and the code book id used for encoding.

%% file: conclusion.tex
This paper addresses the bottleneck of network bandwidth in training and serving Large Language Models (LLMs), where collective operations often limit performance. Traditional Huffman coding provides optimal lossless compression, however, its three-stage process, requiring on-the-fly frequency analysis and codebook transmission introduces latency overheads that are prohibitive for extremely latency sensitive scenarios like die-to-die communication.

We analyzed the FFN1 activation tensors of the Gemma 2B model across 1152 shards. We demonstrated statistical similarity i.e., the histograms of different shards are very similar. A low KL divergence of the individual shards from the average distribution confirmed that the average distribution is a good approximation of the true distribution.
Using a fixed codebook derived from the average distribution achieves a compressibility within 0.5\% of per-shard Huffman codes and within 1\% of the ideal Shannon compressibility.

The codebooks can be pre-computed off the critical path using data from previous batches or runs. We maintain distinct codebooks for different tensors and datatypes. The codebooks are shared between the participating nodes, thereby eliminating the need to transmit them during operation. This approach allows a single-stage compression without incurring the computational and latency overheads of traditional methods.